\pdfoutput=1

\documentclass[11pt]{article}

\usepackage{acl}

\usepackage{times}
\usepackage{latexsym}

\usepackage[T1]{fontenc}

\usepackage[utf8]{inputenc}

\usepackage{microtype}

\usepackage{graphicx}
\usepackage{stfloats}
\usepackage{bm}
\usepackage{amsmath}
\usepackage{array}
\usepackage{booktabs}
\usepackage{multirow}
\usepackage{makecell}
\usepackage{CJKutf8}

\DeclareMathOperator*{\argmax}{arg\,max}

\newcommand{\PreserveBackslash}[1]{\let\temp=\\#1\let\\=\temp}
\newcolumntype{C}[1]{>{\PreserveBackslash\centering}p{#1}}
\newcolumntype{R}[1]{>{\PreserveBackslash\raggedleft}p{#1}}
\newcolumntype{L}[1]{>{\PreserveBackslash\raggedright}p{#1}}

\newcommand{\chinese}[1]{\begin{CJK}{UTF8}{gkai}{}#1\end{CJK}}

\title{LitMind Dictionary: An Open-Source Online Dictionary}

\author{Cunliang Kong\textsuperscript{\dag},
	Xuezhi Fang\textsuperscript{\dag},
	Liner Yang\textsuperscript{\dag}\thanks{\ \ Indicates corresponding author},
	Yun Chen\textsuperscript{\ddag},
		Erhong Yang\textsuperscript{\dag} \\
	\textsuperscript{\dag}School of Information Science, Beijing Language and Culture University \\
	\textsuperscript{\ddag}School of Information Management \& Engineering, \\Shanghai University of Finance and Economics \\
	cunliang.kong@outlook.com, jasonfang3900@gmail.com, 
	yangtianlin@blcu.edu.cn,\\
yunchen@sufe.edu.cn,
	 yerhong@blcu.edu.cn
}

\date{}

\begin{document}
	\maketitle
	
	\begin{abstract}
		Dictionaries can help language learners to learn vocabulary by providing definitions of words.
		Since traditional dictionaries present word senses as discrete items in predefined inventories, they fall short of flexibility, which is required in providing specific meanings of words in particular contexts.
		In this paper, we introduce the LitMind Dictionary (\url{https://dictionary.litmind.ink}), an open-source online generative dictionary that takes a word and context containing the word as input and automatically generates a definition as output.
		Incorporating state-of-the-art definition generation models, it supports not only Chinese and English, but also Chinese-English cross-lingual queries.
		Moreover, it has a user-friendly front-end design that can help users understand the query words quickly and easily.
		All the code and data are available at \url{https://github.com/blcuicall/litmind-dictionary}.
	\end{abstract}
	
	\section{Introduction}
	Helping language learners understand words in doubt is an important topic in the field of Intelligent Computer-Assisted Language Learning (ICALL) \citep{segler-2002-second, enayati-2020-impact, lolita-2020-impact}.
	Most dictionaries present word senses as discrete items in predefined inventories to help language learners understand new words.
	Nevertheless, this form is suffering from several limitations and may bring users inconveniences in many cases.
	First, many commonly used words are polysemous, and it's difficult for language learners to distinguish different word senses because of the cognitively inaccurate nature of discrete sense boundaries \citep{rosch-1975-family,kilgarriff-1997-don,tyler-2001-reconsidering}.
	As \citet{kilgarriff-2007-word} argued, there are no decisive ways of identifying where one sense of a word ends and the next begins.
	In addition, the predefined inventories need to be updated manually by lexicographers, which is time-consuming and causes dictionaries to lag behind the ever-changing language usage.
	For example, many new words have emerged along with the change in people's lifestyles, such as \textit{tweet} (make a post on Twitter) and \textit{geekery} (enthusiasm for a subject).
	However, these words and senses didn't appear in traditional dictionaries until they were used for a long time.
	
	\begin{figure}[t]
		\centering
		\includegraphics[width=\linewidth]{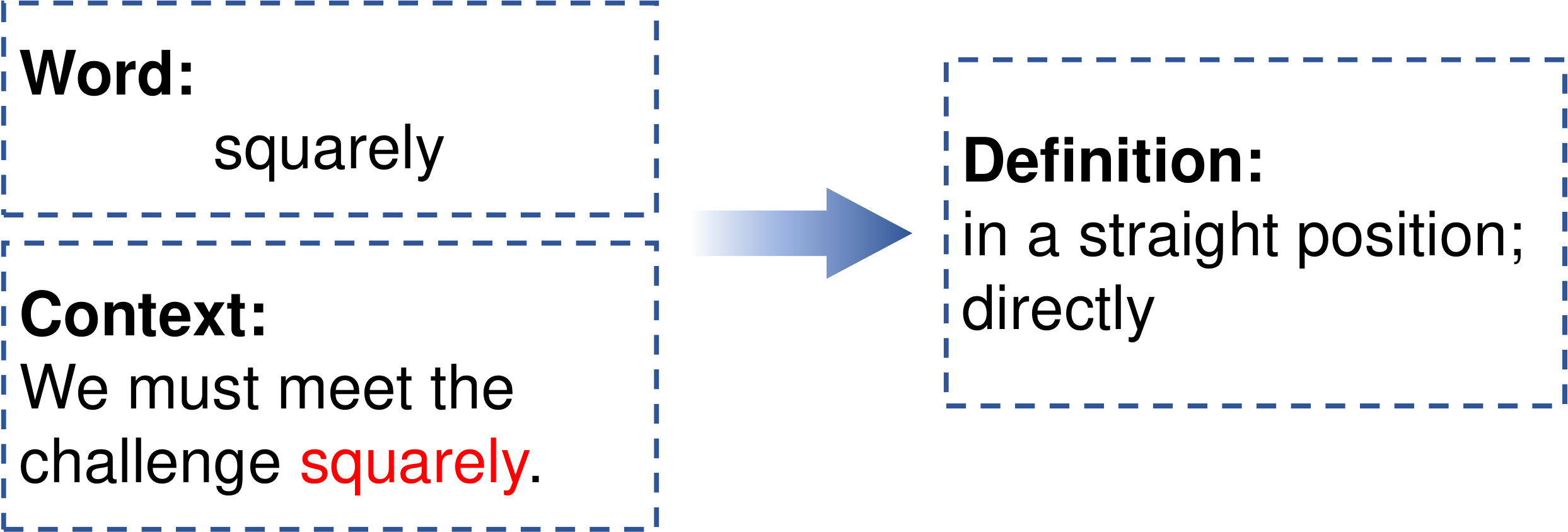}
		\caption{The main functionality and architecture of LitMind Dictionary.}
		\label{fig:arch}
	\end{figure}
	
	We overcome these limitations by developing LitMind Dictionary,\footnote{LitMind is derived from \textit{The \textbf{Lit}erary \textbf{Mind} and the Carving of Dragons}, which is the first systematic literary theory work in China, written around 502 BC.} an open-source online generative dictionary.
	It takes a word and the context containing the word as input and provides automatically generated definitions for the word.
	In this way, the word's definition in current context can be given directly, saving users from selecting from a variety of senses.
	This approach breaks away from the limitation of predefined inventories, with potential to generate correct definitions of new words according to the contexts.
	
	In LitMind Dictionary, the context-aware definitions are generated using a range of NLP and machine learning techniques mainly based on our previous work of definition modeling \citep{yang-2020-incor,fan-etal-2020-ji,kong-2020-toward}.
	The proposed dictionary uses improved versions of these models, and incorporates some engineering tricks to handle extreme cases.
	Moreover, the dictionary supports Chinese and English queries as well as Chinese-English cross-lingual queries, all of which are realized in a generative fashion for the first time.
	Finally, the user-friendly interface design can help users understand the query words as quickly and easily as possible.
	
	\section{Related Work}
	Our work is related to the task of definition modeling, which aims at automatic definition generation.
	This task is first introduced by \citet{noraset-2017-definition}. They used word embeddings as encoded information and a two-layer LSTM as the decoder.
	
	However, their method failed to account for polysemous words, and subsequent work proposed different solutions for this problem.
	\citet{gadetsky-2018-conditional} released a dataset containing example sentences and computed the AdaGram vector \citep{bartunov-2016-breaking} of input words, which is a non-parametric Bayesian extension of skip-gram capable to learn numbers of representations at desired semantic resolution.
	\citet{chang-2018-xsense} proposed to project the given words to high-dimensional sparse vectors, and picked different dimensions for different meanings.
	\citet{mickus-etal-2019-mark} implemented a self-attention based model, and proposed several masking strategies for the input words and example sentences.
	\citet{li-etal-2020-explicit} explicitly decomposed the meaning of words into semantic components, and modeled them with discrete latent variables.
	\citet{yang-2020-incor} explored definition modeling for Chinese and incorporated \textit{sememes} \citep{dong-2006-hownet}, minimal semantic units, as part of the representation of given words.
	\citet{zheng-etal-2021-decompose} proposed to enhance the definition generation with word formation features in parataxis languages like Chinese.
	Besides, \citet{ishiwatari-etal-2019-learning} extended this task to describe unknown phrases.
	They replaced the target word in context with a placeholder, and used a character-level CNN together with static embedding for word representation.
	
	Recent years have witnessed the application of pretrained language models in definition modeling \citep{chang-2019-what}.
	\citet{reid-etal-2020-vcdm} initialized encoders with BERT \citep{devlin-etal-2019-bert} and employed variational inference for estimation and leverage contextualized word embeddings for improved performance.
	\citet{bevilacqua-etal-2020-generationary} employed a novel span-based encoding scheme to fine-tune a pre-trained English Encoder-Decoder system to generate definitions.
	\citet{huang-etal-2021-definition} leveraged the T5 model \cite{raffel-2019-explor} for this task and introduced a re-ranking mechanism to model specificity in definitions.
	
	Our approach follows previous work in using pretrained language models for two reasons: (1) due to the data scarcity in the definition modeling task, it's difficult for other models to obtain better performance; (2) using contextual word embeddings can solve the ambiguity of polysemous words.
	Moreover, we employ cross-lingual pre-trained language models \citep{lample-2019-cross, conneau-2019-unsupervised} as an extension of the task to support Chinese-English queries.
	
	\begin{figure}[t]
		\centering
		\includegraphics[width=\linewidth]{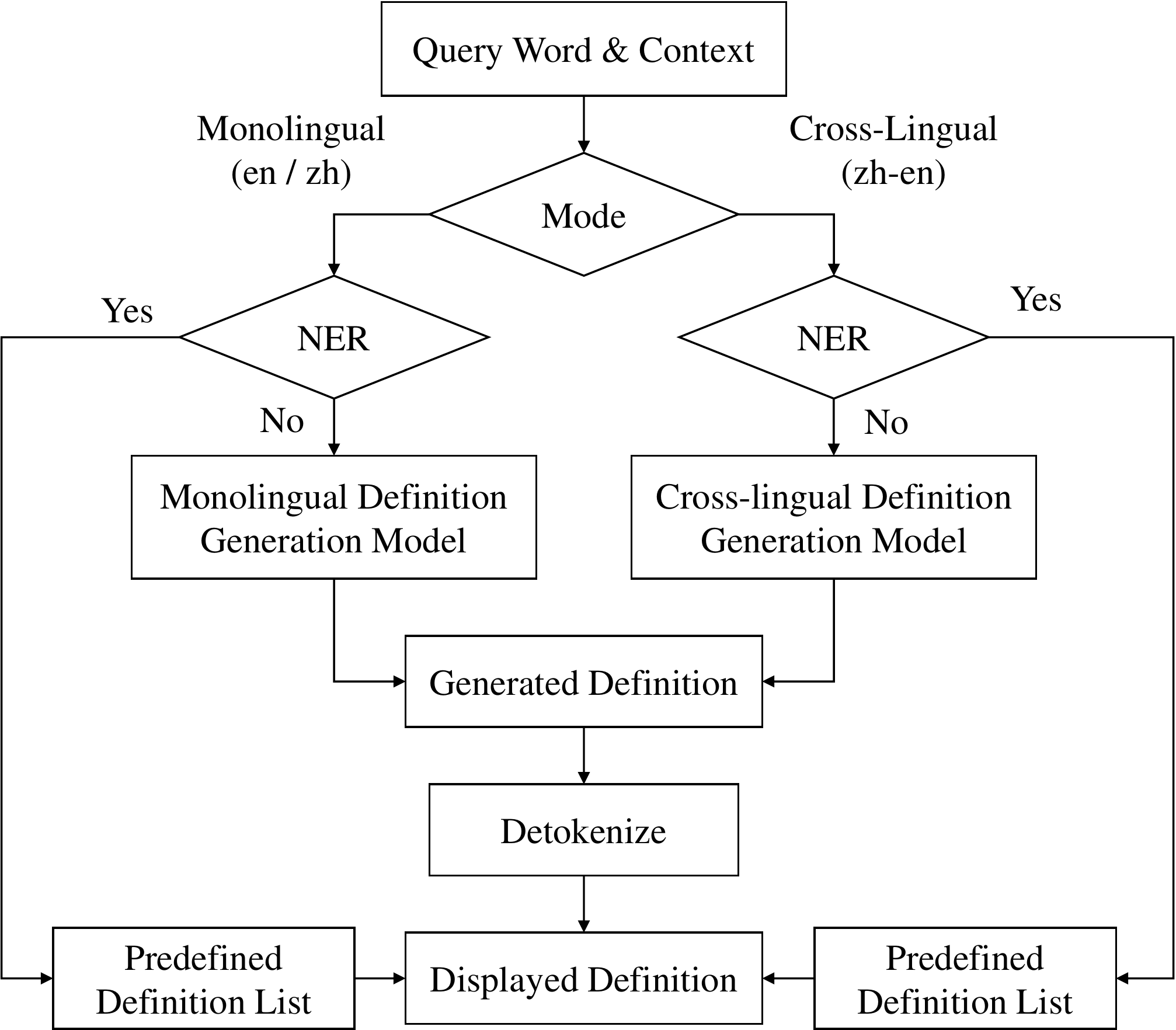}
		\caption{The overall workflow of LitMind Dictionary.}
		\label{fig:workflow}
	\end{figure}
	
	\section{System Description}
	Figure \ref{fig:arch} shows an overview of the main functionality and architecture of LitMind Dictionary. In this section, we first describe its overall workflow, then we detail the definition generation model, and finally we introduce its user interface design.
	
	\subsection{Overall Workflow}
	The workflow of LitMind Dictionary is illustrated in Figure \ref{fig:workflow}.
	When using the dictionary, a user needs to provide the query word and context.
	We require the word to appear in the context.
	If the query word refers to a named entity, the dictionary will directly retrieve the corresponding definition from a predefined list, such as \textit{Name}, \textit{State or Province}, \textit{Organization}, etc.
	Otherwise, the dictionary will feed the query word and context into the monolingual or cross-lingual definition generation model, depending on the user's selection.
	
	\begin{figure}[t]
		\centering
		\includegraphics[width=\linewidth]{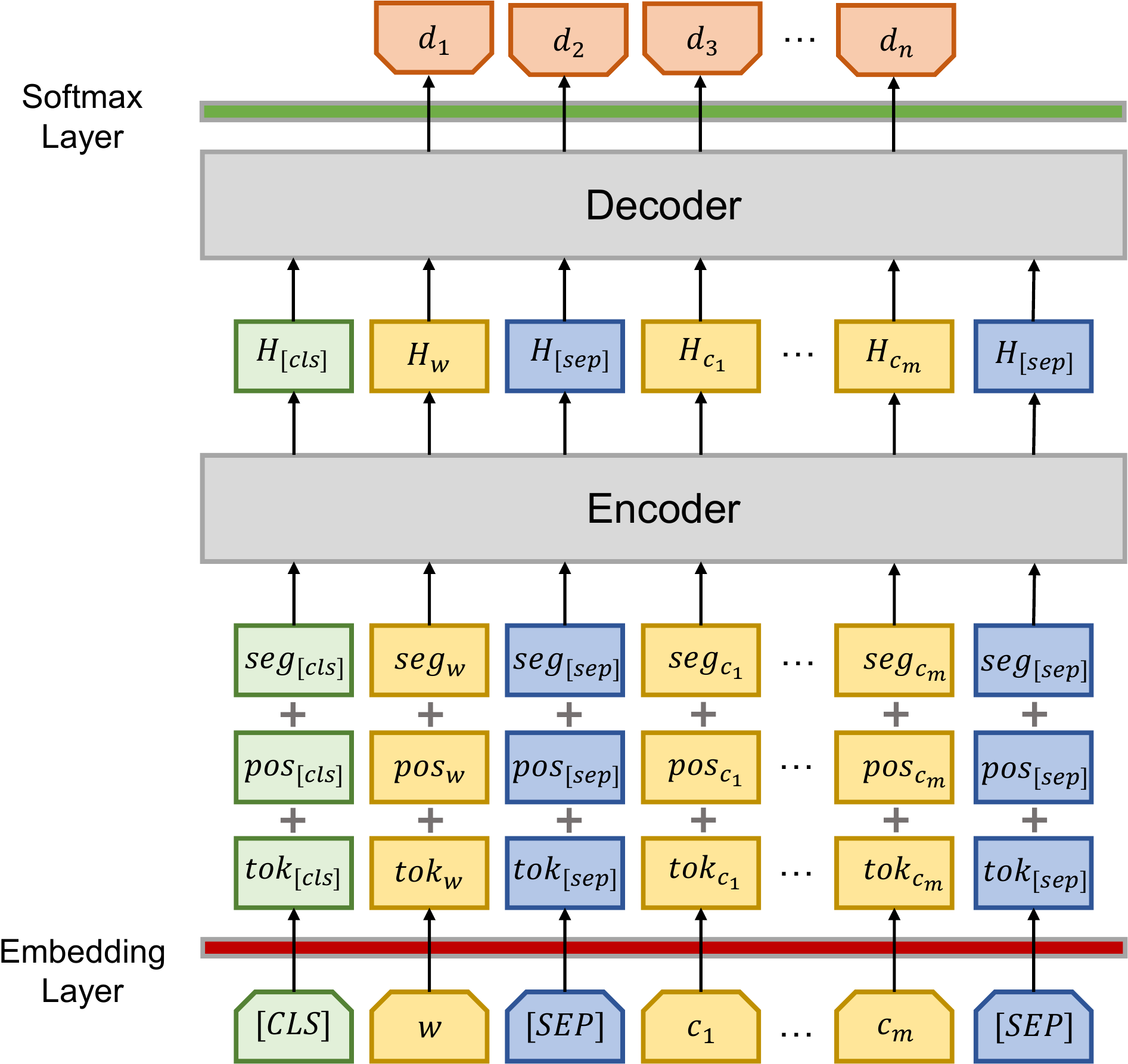}
		\caption{Overall structure of the definition generation model.}
		\label{fig:model}
	\end{figure}
	
	\subsection{Definition Generation Model} \label{section:model}
	The definition generation model (DGM) shown in Figure \ref{fig:model} is the core component of LitMind Dictionary.
	It is used to automatically generate the definition of a given word.
	The whole model is a transformer \citep{vaswani-2017-attention} based encoder-decoder model, and we use pretrained language models to initialize the parameters in experiments.
	
	For a given word $w$ and its corresponding context $\bm{c}$, we segment them into subwords and use each subword as a token.
	To facilitate the word and context encoding, we concatenate then into a whole sequence $(w,$ [SEP]$, \bm{c})$.
	In order to represent the position of different tokens, we add a positional embedding and a segment embedding to each token as \citet{vaswani-2017-attention}'s original setting.
	For the $i$-th token in the sequence, we add its token embedding $tok_i$, position embedding $pos_i$, and segment embedding $seg_i$ together as $\mathbf{x}_i = tok_i + pos_i + seg_i$.
	We input the obtained embeddings into the definition generation model, and train the parameters by minimizing the cross-entropy loss between the generated definitions and the true distributions underlying the dataset:
	\begin{equation}
	\bm{\theta^*} = \argmax_{\bm \theta}\sum_{\bm d \in D}
	\log({P(\bm{d} | \mathbf{x}, \bm \theta)}),
	\end{equation}
	where $\bm{\theta}$ refers to all trainable parameters, and $\bm d$ is the predicted definition.
	
	\subsection{Monolingual Mode}
	LitMind Dictionary supports English and Chinese monolingual modes \citep{yang-2020-incor, fan-etal-2020-ji}.
	Monolingual refers to that the language of the definition is the same as that of the word.
	For English mode, users input English word and context, and our system will return English definitions ($<w^{en}, \bm{c}^{en}> \rightarrow \bm{d}^{en}$).
	
	
	\subsection{Cross-lingual Mode}
	Now we introduce LitMind Dictionary in the cross-lingual scenario.
	The only difference from the monolingual mode is that the language of the input word and context is different from that of the output definition \citep{kong-2020-toward}.
	In Chinese-English mode, the input word and context are in Chinese, and the output definition is in English ($<w^{zh},\bm{c}^{zh}> \rightarrow \bm{d}^{en}$).
	
	
	As high quality Chinese-English dictionary resources are difficult to obtain, we don't train the model on Chinese-English parallel dataset.
	Instead, we only train the model on English-English dataset, and then directly transfer the model to Chinese-English scenario in a zero-shot manner.
	Since multilingual PLMs is capable of encoding sequences in various languages, this zero-shot method shows effective results in our manual evaluation.
	In addition, this approach can be extended to other low-resource languages to help more language learners.
	
	\begin{figure*}[t]
		\centering
		\includegraphics[width=\linewidth]{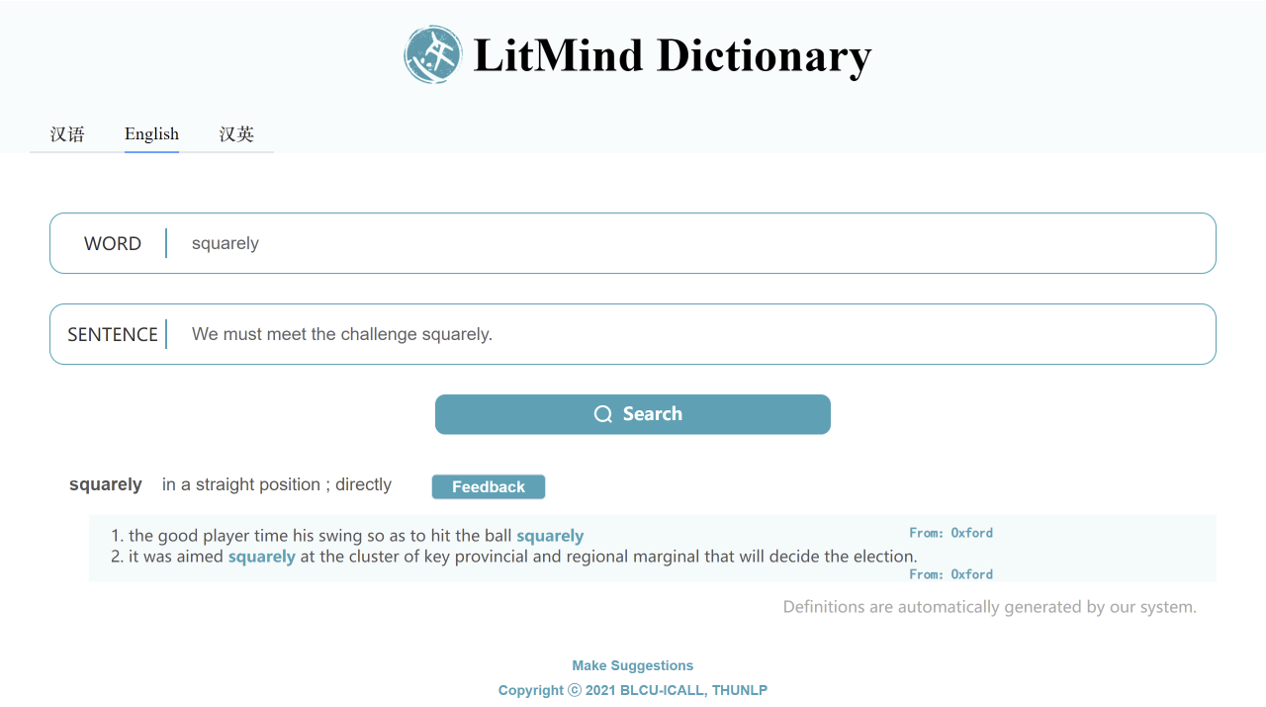}
		\caption{User interface of LitMind Dictionary in the English monolingual mode.}
		\label{fig:interface}
	\end{figure*}
	
	\subsection{User Interface}
	As shown in figure \ref{fig:interface}, the interface of LitMind Dictionary is friendly designed and very easy to use.
	To make a query, users need to input a word and a sentence into the textboxes, and then click the \textit{search} button.
	Our system will automatically generate and display the corresponding definition.
	Under the definition, some example sentences are listed to help users better understand the query word.
	
	We also designed a feedback channel to collect real-world data.
	Users can click the \textit{feedback} button to write the definition they think is appropriate.
	Besides, if users have overall suggestions for LitMind Dictionary, they can click the \textit{Make Suggestions} button to give their advice.
	
	\section{Evaluation}
	In this section, we evaluate the performance of LitMind Dictionary. We conduct both monolingual (Chinese and English) and cross-lingual(Chinese-English) evaluations.
	
	\subsection{Datasets}
	For English monolingual experiments, we use the Oxford dataset published by \citet{gadetsky-2018-conditional} as training, validation and test dataset.
	The dataset has 97,855 entries in the training set, 12,232 entries in the validation set and test set respectively.
	
	As for Chinese, we build a dataset from the Contemporary Chinese Learner's Dictionary (CCLD),\footnote{CCLD: \url{https://www.cp.com.cn/book/9554d669-7.html}} which is specially designed for Chinese learners.
	We first convert each page of the book into text using the optical character recognition (OCR) technology.
	We then recruit a group of annotators to proofread the text and correct the errors generated in the conversion process.
	Finally, we structure the text into the \textit{json} format and extract words, example sentences and definitions and obtain the entire dataset of 6,284 words, 89,065 entries.
	The dataset is then split into three subsets as training set, validation set, and test set by the ratio of 8:1:1.
	
	For the Chinese-English cross-lingual definition generation, we train the model using the above mentioned Oxford dataset.
	For evaluation, we randomly sampled 200 entries from the CCLD test set, and perform zero-shot generation on it.
	Since there are no golden English definitions in the test set, we organize manual evaluation to score the results generated by models.
	
	Tabel \ref{table:data} lists more detailed statistics of the above datasets.
	
	\begin{table}[t]
		\centering
		\begin{tabular}{lR{1.2cm}R{1.2cm}R{1cm}R{1cm}}
			\Xhline{1pt}
			Dataset & Words & Entries & Exp. & Def. \\
			\hline
			Oxford & & & & \\
			\hline
			Train & 33,128 & 97,855 & 17.74 & 11.02 \\
			Valid & 8,867 & 12,232 & 17.80 & 10.99 \\
			Test & 8,850 & 12,232 & 17.56 & 10.95 \\
			\hline
			\multicolumn{5}{l}{CCLD-Mono} \\
			\hline
			Train & 5,028 & 71,328 & 7.06 & 13.41 \\
			Valid & 628 & 8,700 & 6.89 & 13.43 \\
			Test & 628 & 9,037 & 7.37 & 13.47 \\
			\hline
			\multicolumn{5}{l}{CCLD-Cross} \\
			\hline
			Test & 163 & 200 & 7.19 & 14.38 \\
			\Xhline{0.8pt}
		\end{tabular}
		\caption{Statistics of the Oxford dataset, CCLD (Monolingual) dataset, and CCLD (Cross-lingual) dataset. The columns are the number of words and entries, the average length of example sentences and definitions.}
		\label{table:data}
	\end{table}
	
	\subsection{Models}
	In this work, we mainly compare models of three different categories: non-pretrained model, pretrained masked LM, and pretrained encoder-decoder.
	We then present the detailed settings of these models.
	
	\paragraph{Non-Pretrained Model}
	We choose the \textbf{LOG-CaD} \cite{ishiwatari-etal-2019-learning} model as a baseline for monolingual experiments.
	This model is an encoder-decoder model proposed to describe unknown phrases, which can also be used to generate definitions for given words and contexts.
	The model has three different encoders, which are (1) a global context encoder to lookup pretrained embeddings of the given word; (2) a local context encoder (Bi-LSTM) to encode the context; (3) a CNN layer to encode character-level features of the given word.
	The decoder is a two-layer LSTM, which receives the above three encoded information and dynamically weighs them at each time step.
	We set the hyper-parameters exactly like the original paper for a fair comparison.
	
	\begin{table}[t]
		\centering
		\begin{tabular}{l|C{0.9cm}C{0.9cm}|C{0.9cm}C{0.9cm}}
			\Xhline{1pt}
			\multirow{2}{*}{} & \multicolumn{2}{c|}{English} & \multicolumn{2}{c}{Chinese} \\
			\cline{2-5}
			& BLEU & NIST & BLEU & NIST \\
			\hline
			LOG-CaD & 18.76 & 43.97 & 20.71 & 44.57 \\
			BERT-Base & 22.84 & 65.24 & \textbf{23.64} & 41.91 \\
			BERT-Large & 27.35 & 87.32 & -- & -- \\
			(m)BART & \textbf{29.51} & \textbf{98.92} & 22.36 & \textbf{49.53} \\
			\Xhline{0.8pt}
		\end{tabular}
		\caption{BLEU and NIST scores of monolingual experiments. Note that BART is used for English and mBART is used for Chinese.}
		\label{table:mono-bleu}
	\end{table}
		
	\paragraph{Pretrained Masked LM}
	The masked LMs are pretrained by the MLM \cite{devlin-etal-2019-bert} task, which aims to predict masked text pieces based on surrounded context.
	We use the masked LMs to initialize parameters in the transformer encoder.
	For monolingual experiments, we compare the \textbf{BERT-Base} and \textbf{BERT-Large} models.
	We only evaluate the effectiveness of English BERT-Large model since no Chinese model available.
	For cross-lingual experiments, we compare the \textbf{mBERT}\footnote{mBERT: \url{https://github.com/google-research/bert/blob/master/multilingual.md}} and \textbf{XLM-R-Large} \cite{conneau-2019-unsupervised} models.
	In practice, we randomly initialize a transformer decoder and feed the output of masked LMs into the cross-attention mechanism.
	The decoder architecture is set the same as \citet{vaswani-2017-attention}.
	We train the entire model in two phases.
	The first phase fix the parameters in the encoder and train the decoder from scratch by the learning rate of 1e-4.
	And the second phase use a smaller learning rate of 1e-5 to fine-tune the entire model.
	We report test results after the fine-tuning phase in Section \ref{section:results}.
	
	\paragraph{Pretrained Encoder-Decoder}
	The encoder-decoder model leverages a left-to-right LM to generate a sentence conditioned on a separate encoder for a given text.
	We fine-tune the \textbf{BART} \cite{lewis-etal-2020-bart} model on the Oxford dataset for English experiments.
	Since no Chinese BART model available, we fine-tune the \textbf{mBART} \cite{liu-2020-multilingual} model on the CCLD-Mono dataset and set both input and output language prompts as \textit{zh\_CN}.
	For cross-lingual experiments, we fine-tune the \textbf{mBART} model on the Oxford dataset and set both input and output language prompts as \textit{en\_XX}.
	And then alter the input language prompt as \textit{zh\_CN} for cross-lingual inference.
	The fine-tuning learning rate is set to 1e-4.
	
	\begin{table}[t]
		\centering
		\begin{tabular}{l|lrrr|r}
			\Xhline{1pt}
			& & \#1 & \#2 & \#3 & Avg. \\
			\hline
			\multirow{4}{*}{Acc.} & XLM-R-L & 1.10 & 1.04 & 1.12 & 1.09 \\
			& mBART & 2.39 & 1.77 & 2.28 & 2.15 \\
			& mBERT & 2.59 & 2.20 & 2.90 & \textbf{2.56} \\
			\hline
			\multirow{4}{*}{Flu.} & XLM-R-L & 3.39 & 4.27 & 4.83 & 4.16 \\
			& mBART & 4.01 & 4.53 & 4.95 & \textbf{4.50} \\
			& mBERT & 3.93 & 4.50 & 4.80 & 4.41 \\
			\Xhline{0.8pt}
		\end{tabular}
		\caption{Manually evaluated accuracy and fluency scores of the cross-lingual dictionary and baseline models. }
		\label{table:cross-manual}
	\end{table}
	
	\subsection{Evaluation}
	For monolingual methods, we use the BLEU \cite{papineni-2002-bleu} and NIST \cite{doddington-2002-automatic} as automatic evaluation metrics.
	NIST focuses on content words by giving more weightage to them.
	This makes NIST more informative than solely assigning an equal weight to each \textit{n}-grams as BLEU \cite{huang-etal-2021-definition}.
	
	For cross-lingual methods, since there are no golden standard English definitions, we let 3 scorers to manually evaluate the generated results.
	We randomly shuffle a total of 600 definitions generated by 3 models and let scorers rate them independently.
	Specifically, each scorer evaluates a definition on two criteria of accuracy and fluency.
	Both criteria range from 1 to 5, with 1 being the lowest and 5 being the highest.

	\begin{table*}[!htb]
		\centering
		\small
		\begin{tabular}{L{2cm}|L{11cm}}
			\Xhline{1pt}
			\multicolumn{2}{c}{English Monolingual Mode} \\
			\hline
			Word & prominence\\
			\hline
			Context & By the close of the 1870s, Homer had achieved national {\color{red} prominence}.\\
			\hline
			Reference & the state of being important, famous, or noticeable \\
			\hline
			LOG-CaD & the state of being protuberant or [unk] \\
			BERT-Base & the state or fact of being prominent \\
			BERT-Large & the quality or state of being recognized or prominent \\
			\textbf{BART} & the state or fact of surpassing all others; superiority in rank or status \\
			\hline
			\multicolumn{2}{c}{Chinese Monolingual Mode} \\
			\hline
			Word & \chinese{惦记} (thinking about) \\
			\hline
			\multirow{3}{2cm}{Context} & \multirow{3}{13cm}{\chinese{我在这边过得很好，您别总{\color{red}惦记}着。} \\
				(I have a good time here, don't worry about it all the time.)} \\
			& \\
			& \\
			\hline
			\multirow{2}{2cm}{Reference} &
			\multirow{2}{13cm}{\chinese{因想念、担心、期盼等而心里老想着}\\
				(Always thinking about sth. because of missing, worrying, looking forward to, etc.)} \\
			& \\
			\hline
			\multirow{2}{2cm}{LOG-CaD} & \multirow{2}{13cm}{\chinese{用在名词或作品上}\\ (Used in nouns or works.)} \\
			& \\
			\multirow{2}{2cm}{BERT-Base} & \multirow{2}{13cm}{\chinese{无法回忆起曾经发生过的事情；过去知道的事不记得了}\\ (Unable to recall what happened; don't remember old things in the past.)} \\
			& \\
			\multirow{2}{2cm}{\textbf{mBART}} & \multirow{2}{13cm}{\chinese{心里记挂着某人或某事}\\ (Keep thinking about sb. or sth.)} \\
			& \\
			\hline
			\multicolumn{2}{c}{Chinese-English Cross-lingual Mode} \\
			\hline
			Word & \chinese{基础} (foundation)\\
			\hline
			\multirow{3}{2cm}{Context} & \multirow{3}{12cm}{\chinese{楼越高，{\color{red}基础}越要坚实。} \\
				(The higher the building, the more solid the foundation.)} \\
			& \\
			& \\
			\hline
			\multirow{2}{2cm}{Reference} & \multirow{2}{13cm}{\chinese{地面以下用来支撑建筑物的部分}\\ (The part below the ground that supports a building.)} \\
			& \\
			\hline
			XLM-R-L & (with reference to the temperature) free from normal temperatures \\
			mBART & the basic level of difficulty \\
			\textbf{mBERT} & the base of something \\
			\Xhline{0.8pt}
		\end{tabular}
		\caption{Definition generation cases in three different modes.}
		\label{table:samples}
	\end{table*}
	
	\subsection{Results} \label{section:results}
	Table \ref{table:mono-bleu} illustrates results of monolingual experiments.
	We observe that BART performs best among all the models.
	On the English test set, BART yields significantly better results than the other three methods.
	On the Chinese test set, although BART performs slightly worse on BLEU score, it still outperforms other model on NIST score significantly.
	Therefore, the LitMind Dictionary uses BART for both English and Chinese monolingual definition generation.
	
	Table \ref{table:cross-manual} shows the manual evaluation results of the cross-lingual experiments.
	We observe that mBERT get the highest accuracy score and mBART get the highest fluency score.
	We use the mBERT in LitMind Dictionary because: (1) these two models have almost the same level of fluency, and (2) we believe that the accuracy of dictionary definitions are more important than fluency.
	
	\subsection{Case Study}
	Table \ref{table:samples} shows the generated definitions in English, Chinese and Chinese-English modes.
	The models we chose to serve in LitMind Dictionary successfully generate accurate and fluent definitions.
	
	For the English mode, LOG-CaD erroneously uses \textit{protuberant} to explain \textit{prominence}, and generates a special [unk] token.
	Both BERT-Base and BERT-Large use the given word in the definitions, and fail to explain the meaning.
	BART defines the given word as \textit{surpassing} and \textit{superiority}, which is more close to the reference semantically.
	
	For the Chinese mode, LOG-CaD generates a completely irrelevant sentence and fails to explain the given word.
	Bert-Base generates the definition of \textit{\chinese{忘记} (forget)} rather than \textit{\chinese{惦记} (thinking about)}, which basically have the opposite meanings.
	The mBART generates the most relevant definition compared to other models. 
	
	For the Cross-lingual mode, \textit{temperature} generated by XLM-R-L has nothing to do with the given word \textit{foundation}.
	The mBART generate a keyword of \textit{basic}, but \textit{difficulty} also fails to explain the meaning of given word.
	In contrast, the definition generated by mBERT is the most relevant.
	
	\section{Conclusion and Future Work}
	In this paper, we present LitMind Dictionary, an open-source online generative dictionary, which can generate context-aware definitions of a given word.
	Our system supports Chinese and English monolingual queries as well as Chinese-English cross-lingual queries, all of which are realized in a generative fashion for the first time.
	In the future, we will try to control the difficulty of the generated definitions to make it more suitable for language learners.
	We will also work on how to match more appropriate example sentences for the query words.
	
	\bibliographystyle{acl_natbib}
	\bibliography{acl}

\begin{thebibliography}{32}
\expandafter\ifx\csname natexlab\endcsname\relax\def\natexlab#1{#1}\fi

\bibitem[{Bartunov et~al.(2016)Bartunov, Kondrashkin, Osokin, and
  Vetrov}]{bartunov-2016-breaking}
Sergey Bartunov, Dmitry Kondrashkin, Anton Osokin, and Dmitry Vetrov. 2016.
\newblock \href {http://proceedings.mlr.press/v51/bartunov16.html} {Breaking
  sticks and ambiguities with adaptive skip-gram}.
\newblock In \emph{Proceedings of of the 2016 Conference on Artificial
  Intelligence and Statistics}, pages 130--138.

\bibitem[{Bevilacqua et~al.(2020)Bevilacqua, Maru, and
  Navigli}]{bevilacqua-etal-2020-generationary}
Michele Bevilacqua, Marco Maru, and Roberto Navigli. 2020.
\newblock \href {https://www.aclweb.org/anthology/2020.emnlp-main.585}
  {Generationary or {``}how we went beyond word sense inventories and learned
  to gloss{''}}.
\newblock In \emph{Proceedings of the 2020 Conference on Empirical Methods in
  Natural Language Processing}, pages 7207--7221.

\bibitem[{Chang and Chen(2019)}]{chang-2019-what}
Ting-Yun Chang and Yun-Nung Chen. 2019.
\newblock \href {https://www.aclweb.org/anthology/D19-1627} {What {Does} {This}
  {Word} {Mean}? {Explaining} {Contextualized} {Embeddings} with {Natural}
  {Language} {Definition}}.
\newblock In \emph{Proceedings of the 2019 {Conference} on {Empirical}
  {Methods} in {Natural} {Language} {Processing} and the 9th {International}
  {Joint} {Conference} on {Natural} {Language} {Processing}
  ({EMNLP}-{IJCNLP})}, pages 6064--6070.

\bibitem[{Chang et~al.(2018)Chang, Chi, Tsai, and Chen}]{chang-2018-xsense}
Ting-Yun Chang, Ta-Chung Chi, Shang-Chi Tsai, and Yun-Nung Chen. 2018.
\newblock \href {https://arxiv.org/abs/1809.03348} {xsense: Learning
  sense-separated sparse representations and textual definitions for
  explainable word sense networks}.
\newblock \emph{arXiv preprint arXiv:1809.03348}.

\bibitem[{Conneau et~al.(2019)Conneau, Khandelwal, Goyal, Chaudhary, Wenzek,
  Guzm{\'a}n, Grave, Ott, Zettlemoyer, and
  Stoyanov}]{conneau-2019-unsupervised}
Alexis Conneau, Kartikay Khandelwal, Naman Goyal, Vishrav Chaudhary, Guillaume
  Wenzek, Francisco Guzm{\'a}n, Edouard Grave, Myle Ott, Luke Zettlemoyer, and
  Veselin Stoyanov. 2019.
\newblock \href {https://arxiv.org/abs/1911.02116} {Unsupervised cross-lingual
  representation learning at scale}.
\newblock \emph{arXiv preprint arXiv:1911.02116}.

\bibitem[{Devlin et~al.(2019)Devlin, Chang, Lee, and
  Toutanova}]{devlin-etal-2019-bert}
Jacob Devlin, Ming-Wei Chang, Kenton Lee, and Kristina Toutanova. 2019.
\newblock \href {https://www.aclweb.org/anthology/N19-1423} {{BERT}:
  Pre-training of deep bidirectional transformers for language understanding}.
\newblock In \emph{Proceedings of the 2019 Conference of the North {A}merican
  Chapter of the Association for Computational Linguistics: Human Language
  Technologies}, pages 4171--4186.

\bibitem[{Doddington(2002)}]{doddington-2002-automatic}
George Doddington. 2002.
\newblock Automatic evaluation of machine translation quality using n-gram
  co-occurrence statistics.
\newblock In \emph{Proceedings of the Second International Conference on Human
  Language Technology Research}, page 138–145. Morgan Kaufmann Publishers
  Inc.

\bibitem[{Dong and Dong(2006)}]{dong-2006-hownet}
Zhendong Dong and Qiang Dong. 2006.
\newblock \href
  {https://www.worldscientific.com/doi/abs/10.1142/9789812774675_0001} {Hownet
  and the computation of meaning}.

\bibitem[{Enayati and Gilakjani(2020)}]{enayati-2020-impact}
Fatemeh Enayati and Abbas~Pourhosein Gilakjani. 2020.
\newblock \href {https://ojs.unm.ac.id/ijole/article/view/10560} {The impact of
  computer assisted language learning ({CALL}) on improving intermediate {EFL}
  learners’ vocabulary learning}.
\newblock \emph{International Journal of Language Education}, 4(2):96--112.

\bibitem[{Fan et~al.(2020)Fan, Kong, Yang, and Yang}]{fan-etal-2020-ji}
Qinan Fan, Cunliang Kong, Liner Yang, and Erhong Yang. 2020.
\newblock \href {https://www.aclweb.org/anthology/2020.ccl-1.32} {Chinese
  definition modeling based on {BERT} and beam seach}.
\newblock In \emph{Proceedings of the 19th Chinese National Conference on
  Computational Linguistics}, pages 336--348.

\bibitem[{Gadetsky et~al.(2018)Gadetsky, Yakubovskiy, and
  Vetrov}]{gadetsky-2018-conditional}
Artyom Gadetsky, Ilya Yakubovskiy, and Dmitry~P. Vetrov. 2018.
\newblock \href {https://www.aclweb.org/anthology/P18-2043} {Conditional
  generators of words definitions}.
\newblock In \emph{Proceedings of the 56th Annual Meeting of the Association
  for Computational Linguistics (Volume 2: Short Papers)}, volume
  abs/1806.10090, pages 266--271.

\bibitem[{Huang et~al.(2021)Huang, Kajiwara, and
  Arase}]{huang-etal-2021-definition}
Han Huang, Tomoyuki Kajiwara, and Yuki Arase. 2021.
\newblock \href {https://aclanthology.org/2021.emnlp-main.194} {Definition
  modelling for appropriate specificity}.
\newblock In \emph{Proceedings of the 2021 Conference on Empirical Methods in
  Natural Language Processing}, pages 2499--2509.

\bibitem[{Ishiwatari et~al.(2019)Ishiwatari, Hayashi, Yoshinaga, Neubig, Sato,
  Toyoda, and Kitsuregawa}]{ishiwatari-etal-2019-learning}
Shonosuke Ishiwatari, Hiroaki Hayashi, Naoki Yoshinaga, Graham Neubig, Shoetsu
  Sato, Masashi Toyoda, and Masaru Kitsuregawa. 2019.
\newblock \href {https://www.aclweb.org/anthology/N19-1350} {Learning to
  describe unknown phrases with local and global contexts}.
\newblock In \emph{Proceedings of the 2019 Conference of the North {A}merican
  Chapter of the Association for Computational Linguistics: Human Language
  Technologies}, pages 3467--3476.

\bibitem[{Kilgarriff(1997)}]{kilgarriff-1997-don}
Adam Kilgarriff. 1997.
\newblock \href {https://link.springer.com/article/10.1023/A:1000583911091} {I
  don’t believe in word senses}.
\newblock \emph{Computers and the Humanities}, 31(2):91--113.

\bibitem[{Kilgarriff(2007)}]{kilgarriff-2007-word}
Adam Kilgarriff. 2007.
\newblock \href {https://link.springer.com/chapter/10.1007/978-1-4020-4809-8_2}
  {Word senses}.
\newblock In \emph{Word Sense Disambiguation}, pages 29--46. Springer.

\bibitem[{Kong et~al.(2020)Kong, Yang, Zhang, Fan, Liu, Chen, and
  Yang}]{kong-2020-toward}
Cunliang Kong, Liner Yang, Tianzuo Zhang, Qinan Fan, Zhenghao Liu, Yun Chen,
  and Erhong Yang. 2020.
\newblock \href {https://arxiv.org/abs/2010.05533} {Toward cross-lingual
  definition generation for language learners}.
\newblock \emph{arXiv preprint arXiv:2010.05533}.

\bibitem[{Lample and Conneau(2019)}]{lample-2019-cross}
Guillaume Lample and Alexis Conneau. 2019.
\newblock \href {https://arxiv.org/abs/1901.07291} {Cross-lingual language
  model pretraining}.
\newblock \emph{arXiv preprint arXiv:1901.07291}.

\bibitem[{Lewis et~al.(2020)Lewis, Liu, Goyal, Ghazvininejad, Mohamed, Levy,
  Stoyanov, and Zettlemoyer}]{lewis-etal-2020-bart}
Mike Lewis, Yinhan Liu, Naman Goyal, Marjan Ghazvininejad, Abdelrahman Mohamed,
  Omer Levy, Veselin Stoyanov, and Luke Zettlemoyer. 2020.
\newblock \href {https://aclanthology.org/2020.acl-main.703} {{BART}: Denoising
  sequence-to-sequence pre-training for natural language generation,
  translation, and comprehension}.
\newblock In \emph{Proceedings of the 58th Annual Meeting of the Association
  for Computational Linguistics}.

\bibitem[{Li et~al.(2020)Li, Bao, Huang, Dai, and Chen}]{li-etal-2020-explicit}
Jiahuan Li, Yu~Bao, Shujian Huang, Xinyu Dai, and Jiajun Chen. 2020.
\newblock \href {https://www.aclweb.org/anthology/2020.acl-main.65} {Explicit
  semantic decomposition for definition generation}.
\newblock In \emph{Proceedings of the 58th Annual Meeting of the Association
  for Computational Linguistics}, pages 708--717.

\bibitem[{Liu et~al.(2020)Liu, Gu, Goyal, Li, Edunov, Ghazvininejad, Lewis, and
  Zettlemoyer}]{liu-2020-multilingual}
Yinhan Liu, Jiatao Gu, Naman Goyal, Xian Li, Sergey Edunov, Marjan
  Ghazvininejad, Mike Lewis, and Luke Zettlemoyer. 2020.
\newblock \href
  {https://direct.mit.edu/tacl/article-abstract/doi/10.1162/tacl_a_00343/96484}
  {Multilingual denoising pre-training for neural machine translation}.
\newblock \emph{Transactions of the Association for Computational Linguistics},
  8:726--742.

\bibitem[{Lolita et~al.(2020)Lolita, Boeriswati, and
  Lustyantie}]{lolita-2020-impact}
Yuri Lolita, Endry Boeriswati, and Ninuk Lustyantie. 2020.
\newblock \href
  {https://journal.ipm2kpe.or.id/index.php/LEEA/article/view/1896} {The impact
  of computer assisted language learning ({CALL}) use of english vocabulary
  enhancement}.
\newblock \emph{Linguistic, English Education and Art (LEEA) Journal},
  4(1):206--221.

\bibitem[{Mickus et~al.(2019)Mickus, Paperno, and
  Constant}]{mickus-etal-2019-mark}
Timothee Mickus, Denis Paperno, and Matthieu Constant. 2019.
\newblock \href {https://www.aclweb.org/anthology/W19-6201} {Mark my word: A
  sequence-to-sequence approach to definition modeling}.
\newblock In \emph{Proceedings of the First NLPL Workshop on Deep Learning for
  Natural Language Processing}, pages 1--11.

\bibitem[{Noraset et~al.(2017)Noraset, Liang, Birnbaum, and
  Downey}]{noraset-2017-definition}
Thanapon Noraset, Chen Liang, Lawrence Birnbaum, and Doug Downey. 2017.
\newblock \href
  {https://aaai.org/ocs/index.php/AAAI/AAAI17/paper/download/14827/14211}
  {Definition modeling: Learning to define word embeddings in natural
  language}.
\newblock In \emph{Proceedings of the 2017 AAAI Conference on Artificial
  Intelligence}, volume~31.

\bibitem[{Papineni et~al.(2002)Papineni, Roukos, Ward, and
  Zhu}]{papineni-2002-bleu}
Kishore Papineni, Salim Roukos, Todd Ward, and Wei-Jing Zhu. 2002.
\newblock \href {https://doi.org/10.3115/1073083.1073135} {Bleu: A method for
  automatic evaluation of machine translation}.
\newblock In \emph{Proceedings of the 40th Annual Meeting on Association for
  Computational Linguistics}, page 311–318. Association for Computational
  Linguistics.

\bibitem[{Raffel et~al.(2019)Raffel, Shazeer, Roberts, Lee, Narang, Matena,
  Zhou, Li, and Liu}]{raffel-2019-explor}
Colin Raffel, Noam Shazeer, Adam Roberts, Katherine Lee, Sharan Narang, Michael
  Matena, Yanqi Zhou, Wei Li, and Peter~J. Liu. 2019.
\newblock \href {http://arxiv.org/abs/1910.10683} {Exploring the limits of
  transfer learning with a unified text-to-text transformer}.
\newblock \emph{CoRR}, abs/1910.10683.

\bibitem[{Reid et~al.(2020)Reid, Marrese-Taylor, and
  Matsuo}]{reid-etal-2020-vcdm}
Machel Reid, Edison Marrese-Taylor, and Yutaka Matsuo. 2020.
\newblock \href {https://www.aclweb.org/anthology/2020.emnlp-main.513} {{VCDM}:
  {L}everaging {V}ariational bi-encoding and {D}eep contextualized {W}ord
  {R}epresentations for {I}mproved {D}efinition {M}odeling}.
\newblock In \emph{Proceedings of the 2020 Conference on Empirical Methods in
  Natural Language Processing}, pages 6331--6344.

\bibitem[{Rosch and Mervis(1975)}]{rosch-1975-family}
Eleanor Rosch and Carolyn~B Mervis. 1975.
\newblock \href
  {https://www.sciencedirect.com/science/article/abs/pii/0010028575900249}
  {Family resemblances: Studies in the internal structure of categories}.
\newblock \emph{Cognitive Psychology}, 7(4):573--605.

\bibitem[{Segler et~al.(2002)Segler, Pain, and Sorace}]{segler-2002-second}
Thomas~M Segler, Helen Pain, and Antonella Sorace. 2002.
\newblock \href
  {https://www.tandfonline.com/doi/abs/10.1076/call.15.4.409.8272} {Second
  language vocabulary acquisition and learning strategies in {ICALL}
  environments}.
\newblock \emph{Computer Assisted Language Learning}, 15(4):409--422.

\bibitem[{Tyler and Evans(2001)}]{tyler-2001-reconsidering}
Andrea Tyler and Vyvyan Evans. 2001.
\newblock \href {https://www.jstor.org/stable/3086846?seq=1} {Reconsidering
  prepositional polysemy networks: The case of over}.
\newblock \emph{Language}, pages 724--765.

\bibitem[{Vaswani et~al.(2017)Vaswani, Shazeer, Parmar, Uszkoreit, Jones,
  Gomez, Kaiser, and Polosukhin}]{vaswani-2017-attention}
Ashish Vaswani, Noam Shazeer, Niki Parmar, Jakob Uszkoreit, Llion Jones,
  Aidan~N. Gomez, Lukasz Kaiser, and Illia Polosukhin. 2017.
\newblock \href {http://papers.nips.cc/paper/7181-attention-is-all-you-need}
  {Attention is all you need}.
\newblock In \emph{Proceedings of the 2017 Conference on Neural Information
  Processing Systems}, pages 6000--6010.

\bibitem[{Yang et~al.(2020)Yang, Kong, Chen, Liu, Fan, and
  Yang}]{yang-2020-incor}
Liner Yang, Cunliang Kong, Yun Chen, Yang Liu, Qinan Fan, and Erhong Yang.
  2020.
\newblock \href {https://doi.org/10.1109/TASLP.2020.2987754} {Incorporating
  sememes into chinese definition modeling}.
\newblock \emph{IEEE/ACM Transactions on Audio, Speech, and Language
  Processing}, 28:1669--1677.

\bibitem[{Zheng et~al.(2021)Zheng, Dai, Li, Liu, Sui, Chang, and
  Liu}]{zheng-etal-2021-decompose}
Hua Zheng, Damai Dai, Lei Li, Tianyu Liu, Zhifang Sui, Baobao Chang, and Yang
  Liu. 2021.
\newblock \href {https://aclanthology.org/2021.naacl-main.437} {Decompose, fuse
  and generate: A formation-informed method for {C}hinese definition
  generation}.
\newblock In \emph{Proceedings of the 2021 Conference of the North American
  Chapter of the Association for Computational Linguistics: Human Language
  Technologies}, pages 5524--5531.

\end{thebibliography}
	
\end{document}